\pdfoutput=1
\documentclass[11pt]{article}

\usepackage[]{acl}

\usepackage{times}
\usepackage{latexsym}
\usepackage[T1]{fontenc}
\usepackage[utf8]{inputenc}
\usepackage{microtype}
\usepackage{graphicx}
\usepackage{booktabs}
\usepackage{enumitem}
\usepackage{xurl}
\usepackage{subcaption}
\usepackage{xspace}
\usepackage{todonotes}

\newcommand{\workbench}{\textsc{{RAGaphene}}\xspace}
  
\title{\workbench{}\\ A RAG Annotation Platform with Human ENhancements and Edits}

\author{
    Kshitij Fadnis\thanks{Authors Contributed Equally}, 
    Sara Rosenthal\footnotemark[1],
    Maeda Hanafi,
    Yannis Katsis,
    Marina Danilevsky
    \\
    \texttt{\{kpfadnis,sjrosenthal\}@us.ibm.com}\\
  IBM Research - AI 
  }
  
\begin{document}

\maketitle

\begin{abstract}
Retrieval Augmented Generation (RAG) is an important aspect of conversing with Large Language Models (LLMs) when factually correct information is important. LLMs may provide answers that appear correct, but could contain hallucinated information. Thus, building benchmarks that can evaluate LLMs on multi-turn RAG conversations has become an increasingly important task. Simulating real-world conversations is vital for producing high quality evaluation benchmarks. We present \workbench{}, a chat-based annotation platform that enables annotators to simulate real-world conversations for benchmarking and evaluating LLMs. \workbench{} has been successfully used by approximately 40 annotators to build thousands of real-world conversations. 
\end{abstract}

\section{Introduction}

 \begin{figure}[t]
    \centering
\includegraphics[width=\columnwidth]{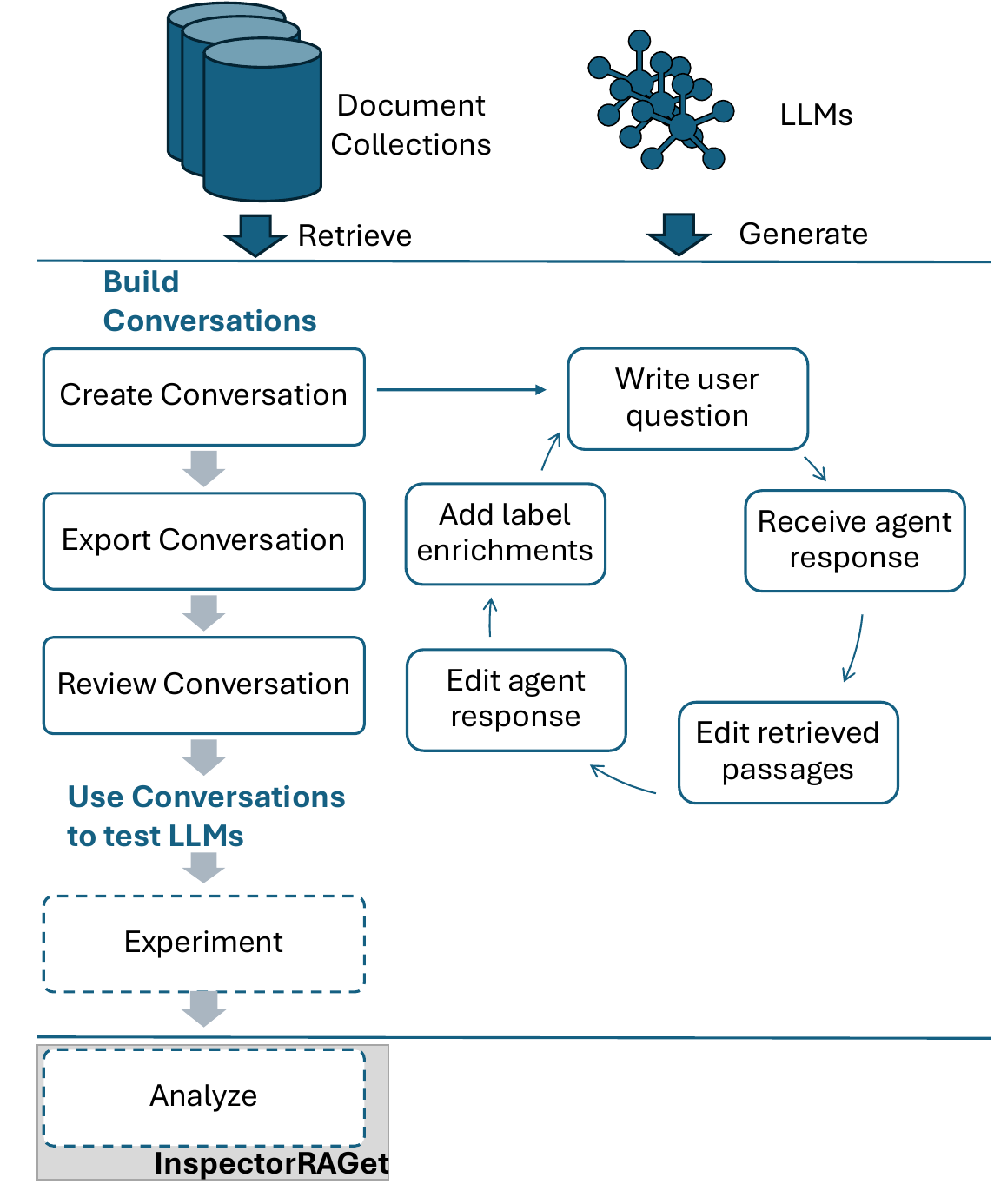}
\caption{The pipeline of \workbench{}: A collection of documents and an LLM are chosen as the desired retriever and generator. The user uses \workbench{} to create a conversation which can be exported in a structured json format for further review and optional experimentation and analysis.}
\label{fig:pipeline}
\end{figure}

\begin{figure*}[t]
    \centering
    \includegraphics[width=\textwidth]
    {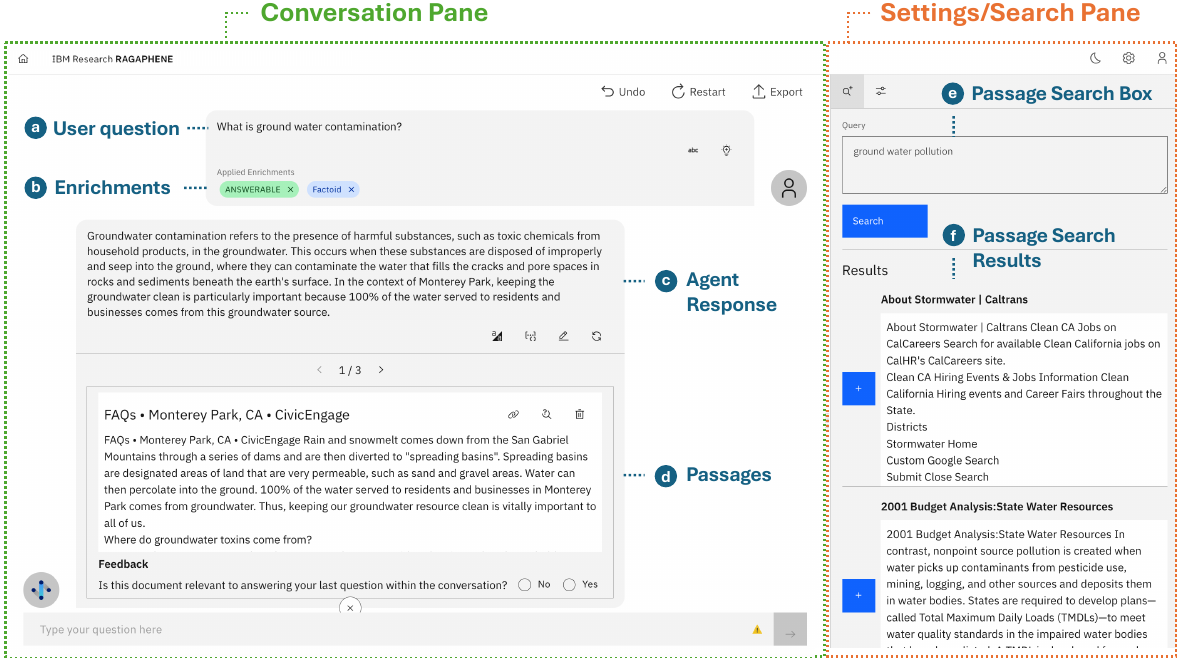}
    \caption{Screenshot of \workbench's create mode annotated with its main components.}
    \label{fig:main-ui}
\end{figure*}

Chat-Based web tools for conversing with Large Language Models (LLMs) such as ChatGPT \cite{OpenAI_ChatGPT} and Claude \cite{claude3} have become extremely popular. One common use is the seeking of factual information where Retrieval Augmented Generation (RAG) is extremely important to ensure the model is answering faithfully to the relevant passages and not hallucinating. Thus, the ability to evaluate the performance of LLMs on multi-turn RAG-based conversations has become increasingly important and largely overlooked until recently \cite{katsis2025mtrag, dziri-etal-2022-faithdial, feng2021multidoc2dial, kuo2024radbenchevaluatinglargelanguage, es-etal-2024-ragas}. Multi-Turn RAG is particularly important for enterprise use cases where domains are specific and there may be unique requirements such as specialized retrievers, generators and custom prompts ~\cite{sharma2024retrievalaugmentedgenerationdomainspecific}. Building a challenging high-quality Multi-Turn RAG benchmark requires in-depth fine-grained human annotation that employs these requirements. Existing annotation platforms \cite{BasicAI, LabelStudio, Labelbox, conventitylinking} include some basic features for annotating conversational data, such as thumbs up/down of questions/responses, adding metadata and tagging entities in the conversation. However, they do not support the creation of conversational multi-turn RAG datasets with real-time agent response generation. The First-Aid platform \cite{menini-etal-2025-first} is a conversational annotation interface which does include real-time agent response generation and editing. However, it doesn't have any other feedback mechanisms (e.g. thumbs up/down) and does not incorporate a retrieval component - rather the conversations are generated based on a small set of pre-loaded documents as opposed to passages dynamically retrieved from a potentially very large underlying corpus. In contrast to prior work, our annotation tool enables editing/correcting both the relevant passages and output which are both important pieces for simulating real-world conversations. 

 We present \workbench{}, a chat-based annotation platform for having conversations with an LLM that are grounded in an existing corpora to ensure faithfulness, primarily for building Multi-turn RAG benchmarks. The pipeline of \workbench{} is shown in Figure~\ref{fig:pipeline}. Our platform adopts the RAG pipeline to enable a user to chat with an LLM agent with real-time retrieval and generation while providing the ability to improve the conversation when the retriever and/or generator fails. We allow the user to easily integrate with their desired retriever and generator and provide the ability to improve the conversation which enables users to create high-quality conversations that can be used to evaluate and improve RAG systems. 


It is desirable for a platform for building and evaluating a challenging multi-turn to have the following: 

\textbf{RAG Chat} RAG-based chat with customizable retrieval and generation is an important scenario, particularly in industry when domain specific corpora is likely. 

\textbf{Enhanced Feedback} Enhanced feedback is desired as typical chat feedback of thumbs up/down is limiting in its use. It does not give the user the opportunity to troubleshoot when the agent response is not satisfactory which is necessary for building high-quality conversations.

\textbf{Evaluation and Analysis} Domain experts and engineers may want the ability to quickly evaluate how different models perform in their domain by building a small benchmark.

Our contributions are as follows:

\begin{itemize}[leftmargin=*,noitemsep]
    \item \workbench{}: A RAG annotation chat platform with integration to many retrievers and generators that can be inter-changed for domain specific use cases. \footnote{We plan to release the \workbench{} code on github following internal approval.}
    \item Enhanced user feedback including adjusting passage retrieval and improving/repairing responses and a user study which highlights the importance of these contributions.
    \item Real-time small-scale evaluation and analysis of the full RAG pipeline on RAG conversations created in \workbench{}. 
\end{itemize}

\section{\workbench{} Platform}

\begin{figure*}[t]
    \centering
    \begin{subfigure}{.26\textwidth}
    \centering
    \includegraphics[width=.95\textwidth]{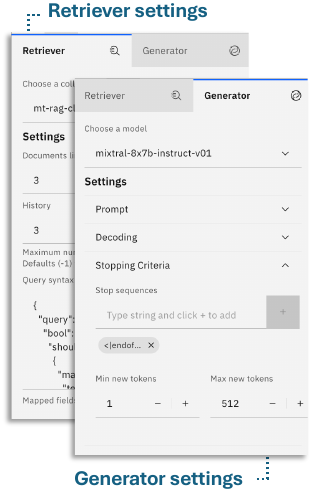}
    \caption{Customizable retriever and generator settings}
    \label{fig:settings-ui}
    \end{subfigure}
    \begin{subfigure}{.35\textwidth}
    \centering
    \includegraphics[width=.95\textwidth]{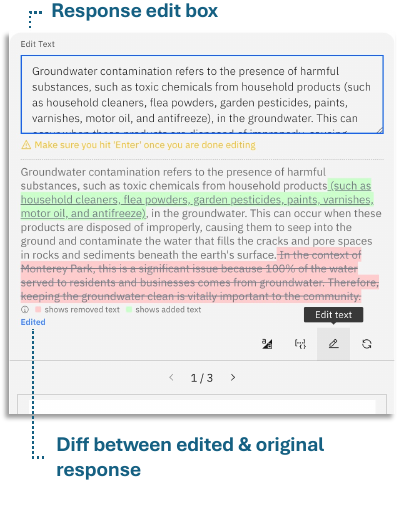}
    \caption{Response editing functionality}
    \label{fig:response-edit-ui}
    \end{subfigure}
    \begin{subfigure}{.35\textwidth}
    \centering
    \includegraphics[width=.95\textwidth]{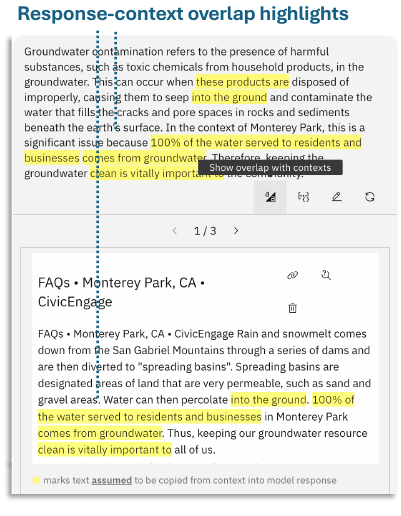}
    \caption{Response-context overlap highlighting}
    \label{fig:response-highlight-ui}
    \end{subfigure}
    \caption{Screenshots of selected functionalities of \workbench.}
\end{figure*}

The \workbench{} platform pipeline is shown in Figure~\ref{fig:pipeline}.  It allows for integration using a desired corpus with any retriever (e.g. BM25, Elser\footnote{\url{https://www.elastic.co/guide/en/machine-learning/current/ml-nlp-elser.html}}) and generator (e.g. Llama 3 \cite{grattafiori2024llama3herdmodels}, GPT 4 \cite{openai2024gpt4technicalreport}). The default settings used in our work are an ELSERv1 (ElasticSearch 8.10\footnotemark[1]) retriever populated with the corpora from MTRAG \cite{katsis2025mtrag} and Mixtral 8X7b Instruct \cite{mixtral} as the generator. \workbench{} consists of three main modes: Create, Review and Experiment as well as integration with InspectorRAGet for analysis to provide a full benchmarking and evaluation life-cycle. We adopt a stateless approach to ensure privacy; all conversations can be preserved by exporting in a structured json format.

\subsection{Create Mode}

In the create mode a user or annotator can use \workbench{} as a chat interface (Figure~\ref{fig:main-ui}) to interact with an LLM. The user is first provided with configuration settings on the right hand side to choose the desired retriever and generator (Figure~\ref{fig:response-highlight-ui}). In the retriever settings, they can pick a specific collection of documents and adjust settings such as how many passages to return, how to formulate the query, and other more sophisticated features such as how to render retrieved results. In the generator settings, they can choose a model to chat with, adjust the prompt and other settings such as decoding and number of tokens. On the left-hand side, the user can chat with the generator chosen in the settings. 
We describe the process of conversation creation at each turn and all of its features including adding question enrichments, answer repair (Figure~\ref{fig:response-edit-ui}), and passage search (Figure~\ref{fig:main-ui}) in Section~\ref{sec:create_flow}. During conversation creation, we also provide tips to assist the user which appear on the upper portion of the interface to encourage useful feedback. The completed conversation can be exported as a structured json for future use.



\subsection{Review Mode}

The review mode is created specifically for cases where \workbench{} is being used as an annotation platform to review previously created conversations and decide whether the conversation should be kept in the benchmark (accept) or not (reject). In this mode, an existing conversation or batch of conversations is loaded and the retriever and generator cannot be run. The annotator can read through the conversation and at each turn modify the enrichments, edit the answers, and change the relevance of the passages. On the left hand side there is a comment section (Appendix, Figure~\ref{fig:comment-ui}) where feedback can be provided for all changes made. Specific comments can be left by highlighting text in the conversation and general comments can be left for overall feedback. Once the conversation review is complete the annotator accepts or rejects the conversation and then continues to the next conversation. When all conversations have been reviewed they can export their work as a structured json file. We describe the flow of the review process in detail in Section ~\ref{sec:review_flow}.

\subsection{Experiment Mode}

The experiment mode provides a way for users to estimate the complexity of their newly created conversational data by running a small scale experiment against various LLMs. In this mode, the prediction by the LLM is compared to the target response that was created during create mode. The user first uploads their conversational data and then chooses which part of the conversation(s) to evaluate. They can choose to split each conversation at every turn, at the last turn, at the beginning of the conversation or at a random turn. Each split becomes a task to be evaluated. Additionally, a user can choose to experiment either with the model's generation capabilities by keeping the retrieved documents constant or to execute the full RAG pipeline using different retriever and generator combinations. The user can quickly adjust the retriever and generator configurations and select evaluation metrics before launching the experiment which can be monitored live (Appendix, Figure~\ref{fig:experiment-ui}). The platform has built-in metrics such as response length, ROUGE, Recall and LLM-as-a-Judge which can be easily extended to run complex compute intensive metrics. Once the experiment is complete, the results can be exported in a structured json format. We describe the flow of the experiment process in detail
in Section~\ref{sec:experiment_flow}. We purposefully restrict the size of the dataset in this mode to a maximum of 100 tasks. \workbench{} is not intended to be a full blown evaluation platform, but rather a quick way of identifying and monitoring data on a small scale. Larger experiments should be run offline.


\subsection{InspectorRAGet}

InspectorRAGet \cite{fadnis2024inspectorragetintrospectionplatformrag} is an introspective platform for debugging and analyzing model output and evaluation metrics. Once a conversation has been created and experiments run, it is important to be able to analyze the performance in a general and detailed manner across multiple models and metrics to enable quick decision making regarding model and metric choice for deployed systems. The output from the experiment mode can be loaded into InspectorRAGet to achieve this.

\section{WB Workflow}

In this section we describe the typical workflow of a user in \workbench{} for the different modes: Create, Review, and Experiment.

\subsection{Conversation Creation}
\label{sec:create_flow}

Our custom chat enables users to chat with a live RAG agent consisting of a retriever and generator and correct the retriever and generator outputs as desired. In many scenarios, we expect the conversation is domain specific and has a pre-decided corpus such as for a specific company or topic. In particular, after customizing the retriever and generator settings (see Figure \ref{fig:settings-ui}), users can use the application to perform the following actions at every turn in the conversation:

\noindent (i) {\bf Write user question:} The conversation begins by writing an initial question for the associated domain (see Figure \ref{fig:main-ui}a). We expect the user to have some basic domain knowledge, but they ask a question without seeing a document. 

\noindent (ii) {\bf Receive agent response:} Once the user writes a question, the agent calls the retriever to retrieve potentially relevant passages and then the passages along with an appropriate prompt are sent to the generator to produce a response which is presented to the user (see Figure \ref{fig:main-ui}c). The passages may not always be relevant and the response may not be appropriate. The next steps provide the user with a means of improving the passages and response to help the user receive a helpful and correct answer.


\noindent (iii) {\bf Edit retrieved passages:} The passages returned by the retriever can be reviewed and edited to generate a set of passages that are indeed relevant to the question (see Figure \ref{fig:main-ui}d). This includes (a) discarding (or marking as irrelevant) passages returned by the retriever that are deemed non-relevant to the question, as well as (b) adding other relevant passages present in the corpus that were missed by the retriever. To facilitate the latter, we provide a separate search interface on the side that allows users to try alternate formulations of their question to bring in additional relevant passages (see Figure \ref{fig:main-ui}e and \ref{fig:main-ui}f). The user can then regenerate the response so that it is based on the latest relevant passages. The set of relevant passages can be used to evaluate and improve the retrieval component of RAG systems. We provide reminder tips if they forget to mark relevant passages.

\noindent (iv) {\bf Edit agent response:} Once the relevant passages are identified, if the response is still not appropriate, users can edit the generated agent response to repair and improve it using the relevant passages. This helps ensure that the next turn in the conversation will be based on the correct information and can be used for future LLM improvement.
As the users edit the response, \workbench shows the diff between the original and edited response (see Figure \ref{fig:response-edit-ui}). Finally, to help users check if the response is faithful to the passages, the platform highlights the lexical overlap between the response and the passages (see Figure \ref{fig:response-highlight-ui}).   

\noindent (v) {\bf Add Label Enrichments:} Each turn can also be enhanced with tags, such as question type (e.g. factoid, opinion), answerability (e.g. answerable, unanswerable), and multi-turn (e.g. clarification, follow up) (see Figure \ref{fig:main-ui}b).  We provide reminder tips to encourage this feedback if they forget to add enrichments. These enrichments provide a valuable way of exploring the data during evaluation. 

\noindent (vi) {\bf Export Conversation:} When the person is finished chatting they can save their conversation by exporting it as a json. This can later be used to continue the conversation, or for evaluation of RAG systems. During export they are provided with optional checkboxes to encourage high quality including providing statistics about the conversation and enrichments that were added (see Figure \ref{fig:export-ui}).


\subsection{Conversation Review}
\label{sec:review_flow}

Creating conversations is a comprehensive task with many parts. In order to use these conversations for evaluation they need to be of high quality with all repairs made. The review mode specifically targets users that are annotators tasked with providing high quality conversations for evaluation. In this mode a reviewer will receive a batch of conversations. The reviewer reads through each conversation and can accept the conversation as is, accept it but edit the conversation, or reject it:

\noindent (i) \textbf{Accept Conversation:} If the conversation is a good conversation that flows well the reviewer can accept the conversation as is. However, in many cases there will be some adjustment needed. 

\noindent (ii) \textbf{Accept with Edits:} Even if the conversation is good, it may still need small tweaks or repairs to improve responses according to desired properties. For example, there may be a sentence in the response that is not faithful to the passages or is misspelled. In addition to changing the response, the reviewer can change passage relevance and adjust the enrichments. A reviewer cannot change the questions or query for more relevant passages as that could alter the conversation significantly causing the initial intent to no longer be valid.

\noindent (iii) \textbf{Reject:} In some cases the reviewer may attempt to repair the conversation but the changes can be too significant. There can be a repeated or unnecessary question that disrupts the conversation flow. It can also be clear that the passages provided are not sufficient. In this case the reviewer can reject the conversation instead of repairing it.

\noindent (iv) \textbf{Feedback:} In addition, the reviewer can leave specific and general comments for the creator of the conversation. They can highlight specific parts/elements of the conversation and leave textual comments (Appendix, Figure \ref{fig:comment-ui}). Feedback is particularly useful when a conversation is rejected.

\subsection{Running Experiments}
\label{sec:experiment_flow}

We provide the experiment mode to give users the ability to run quick experiments on a small scale which can then be exported to InspectorRAGet for quick decision making. In this mode, the user can setup an experiment on a set of conversations and choose the tasks, models and metrics to evaluate. 

\noindent(i) \textbf{Choose Tasks:} The user first uploads the conversation and selects the slice of the data they would like to evaluate. This can be all turns of the conversation or different subsets. 

\noindent(ii) \textbf{Choose Models:} The user can choose to evaluate the retriever, generator, or full RAG pipeline. The user then sets up the retriever and generator model settings for each model they want to evaluate. This includes adjusting parameters and prompts.

\noindent(iii) \textbf{Choose Metrics:} Next, the user chooses the metrics to evaluate on each task. We provide built-in metrics such as Rouge, Recall and LLM judges. 

\noindent(iv) \textbf{Run Experiment:} Once satisfied with the experimental setup, the user launches an experiment. The time to run the experiments depends on the number of models, tasks, and metrics being evaluated. Once completed, the evaluation can be exported and loaded in InspectorRAGet for analysis.

\section{Use Cases}

We describe three specific uses and applications of \workbench{}: Multi-Turn RAG annotation, RAG chat assistant and Real-Time Evaluation.

\subsection{Multi-Turn RAG Annotation}

 The primary use of \workbench{} is to create a challenging benchmark that simulates real world conversations to evaluate the performance of RAG systems. As an annotation task, the various feedback and repair mechanisms provide a valuable resource for creating a strong benchmark that is diverse and challenging for LLMs. While interacting with the RAG-based agent, the annotators add feedback that enriches, repairs, and improves the responses. They then export their conversation so it can be used for evaluation and insights. Creating such conversations is a sophisticated process and prone to errors. We specifically employ the review workflow to ensure the conversations for improvement are of high quality. We have successfully created 110 high quality Multi-Turn RAG conversations using the creation and review workflows which have been released as a public benchmark \cite{katsis2025mtrag}. In addition, the platform is being used for ongoing work in this area, with over 5,000 conversations created and over 1,000 conversations reviewed by over 30 annotators. 

\subsection{RAG Chat Assistant}

Although primarily built as an annotation platform, \workbench{} can also be used  as a RAG-based chat platform by a standard user seeking information. RAGFlow \cite{RagFlow} is an open-source RAG engine which can be used to build a customized chat platform using desired retriever and generator capabilities. On the other hand, \workbench{} has the same capabilities and also provides the ability for the user to repair and redirect the agent by looking for better passages and improving the answer to help the agent answer correctly in later turns.

\subsection{Real-Time Evaluation}

A typical client needs to perform due-diligence prior to selecting a vendor that can power their chat-based assistant over a large collection of proprietary documents. This typically involves exploring multiple LLMs and retriever engines to identify the best setup for their domain. Using \workbench{}, they can create domain specific conversations on their content that can power evaluations, performance analysis (via InspectorRAGet \cite{fadnis2024inspectorragetintrospectionplatformrag}), and go / no-go decisions for stakeholders. We have several clients that use \workbench{} to perform such due-diligence.

\section{User Study}

\begin{table}[tb]
    \small
    \centering
    \begin{tabular}{p{0.6\columnwidth}p{0.17\columnwidth}p{0.05\columnwidth}}
        \toprule
            Platform Feature 
            & $\mu$
            & $\sigma$ \\
        \midrule
        \midrule
            Editing the agent responses
            & 4.26
            & 0.82\\
        \midrule
            Highlights for overlapping text in the context and response 
            & 4.13 
            & 0.88\\
        \midrule
            Regenerating the agent response
            & 4.10
            & 0.98\\
        \midrule
            Requery Tool 
            & 3.97
            & 1.37\\
        \midrule
            Checklist before export
            & 3.74
            & 1.34\\
        \midrule
            Marking contexts relevant/irrelevant
            & 3.26
            & 1.43 \\
        \midrule
            Hints 
            & 3.19   
            & 1.42\\
        \midrule
            Enriching the questions
            & 2.74
            & 1.24\\
        \bottomrule
    \end{tabular}
    \caption{\workbench{} features ranked based on their impact to the quality of created conversational data. A higher average value ($\mu$) indicates a bigger impact (1 is ``No decrease [in data quality] at all'' and 5 is ``Extreme decrease in quality''.)}
    \label{tbl:survey}
\end{table}

We performed a user study to understand the importance of our tool for helping annotators create high-quality conversations for benchmarking RAG. 
We surveyed 31 professional annotators (21 females, 10 males), with 13 of them having more than 3+ years experience of annotating. 
At the time of this writing, the tool has been in production for more than 9 months, with most of the annotators creating and reviewing 75+ conversations. Creating and annotating a conversation for RAG applications is a hard and time-consuming task for human annotators \cite{hanafi2025workbench}; 
Most annotators reported that it took them on average more than 30 minutes to create a single high-quality conversation. 

The survey focused on questions for 8 platform features: ``On a scale of 1 to 5 if we were to remove <FEATURE> from the platform, how would this decrease the quality of the conversations you create?'', where 1 is ``No decrease at all'' and 5 is ``Extreme decrease in quality''. Table \ref{tbl:survey} shows the average Likert scores of each platform feature.

Although our professional annotators reported an advanced beginner-level of RAG understanding (on a 1 to 5 Likert scale; $\mu=2.61, \sigma= 1.33$), our survey results show that they recognized the necessity for platform features that involve analyzing and fixing the generator and retriever model outputs for high-quality RAG data creation (such as improving the set of retrieved contexts or editing the agent response).
Certain features, such as marking contexts as relevant or irrelevant and enriching the questions are seen as annotation artifacts rather than part of the process of \textit{creating} a conversation. One annotator says, ``[Marking contexts as relevant/irrelevant] is perhaps useful to the end-user of the data, but is not necessary for creating the data in the tasks...''.

\section{Conclusion}

We present \workbench{}, a platform for creating and evaluating high quality conversations for RAG. We provide advanced features to improve passage search, repair responses, and enrich the questions to aid in conversation creation. We also include the option of performing small-scale evaluation with integration to InspectorRAGet \cite{fadnis2024inspectorragetintrospectionplatformrag} to help perform due-diligence in model selection for domain specific use. Our user study shows that the provided features, such as highlighting and editing responses, improve the quality of the conversations created in the platform.  We plan to release the \workbench{} code with Apache 2.0 license on GitHub following internal approval.

\section{Ethical Considerations}

We take care to ensure the use of our platform is private by providing a stateless approach. Our platform only requires an email login, but does not store any personal information or retain created conversations. We acknowledge that there are accessibility limitations in the platform and we plan on providing additional features to improve these limitations.

During our user study we asked for some personal details to understand our annotators background. All such details were only looked at in aggregate and not attributed to the annotator individually. 

\bibliography{custom}
\clearpage

\appendix

\section{Tooling}
\workbench{} is a React web application built with NextJS 14 framework\footnote{\url{https://react.dev}, \url{https://nextjs.org}} and requires Python >= 3.10 with minimal dependencies to power the experiment flow. We use the Carbon Design System\footnote{\url{https://carbondesignsystem.com}} for the user interface. \workbench{} has built-in connectivity to ElasticSearch, MongoDB Atlas, IBM Cloudant retrieval engines and WatsonX.AI and OpenAI generator engines via corresponding Node SDKs with support for ChromaDB (vector database for retrieval), Claude from Anthropic and vLLM to follow soon. Furthermore, \workbench{} can be extended to support any RESTful retrieval and generation engines with limited data transformations. Our platform is lightweight; it can easily be run on virtual machines or even personal laptops/desktops with 2 CPUs and 8GB RAM. The bulk of the resource consumption is observed only during an experiment run.

To enable privacy, \workbench{} is a stateless application and does not retain any created, reviewed conversations or uploaded datasets. We have defined a simple json format for storing conversations as explained in Appendix \ref{sec:file_format}.

\section{Conversation File Format}
\label{sec:file_format}
As \workbench{} is a web application, we naturally gravitated towards adopting JSON as the input format. Our prescribed structure for an conversation file is intuitive and strives to minimize repetition of information. 

The conversation file can be broadly split into four sections along their functional boundaries. The first section captures general details about the \emph{participants}, including author, editor, and reviewer emails and timestamp of access. The second section describes the \emph{retriever} and \emph{generator} used during creation. The \emph{retriever} and \emph{generator} sub-sections contain connectivity details, customized parameters and some additional system specific settings. The third section captures messages exchanged between the user and assistant. Each message has \emph{speaker} and \emph{text} information. Additionally, in case of user messages, \emph{enrichments} information is also retained. Similarly for agent messages, the retrieved documents/passages are preserved under \emph{contexts}. Finally, the fourth section includes information about the \emph{status of the conversation}, in the form of current status, previous revisions and any comments made during the reviewing process.

\section{Additional Screenshots}

\begin{figure}[t]
    \centering
    \includegraphics[width=.65\columnwidth]{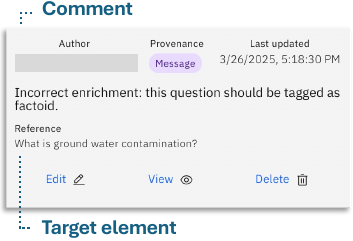}
    \caption{Screenshot of review feedback functionality.}
    \label{fig:comment-ui}
\end{figure}

\begin{figure}[t]
    \centering
    \includegraphics[width=.99\columnwidth]{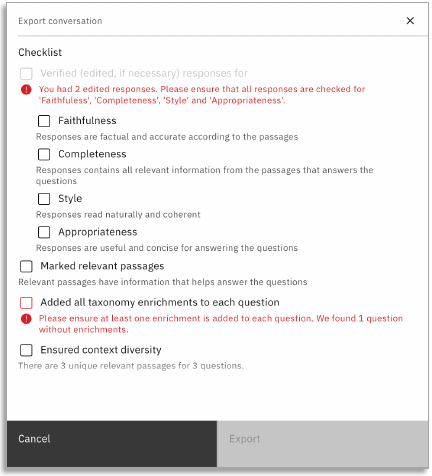}
    \caption{Screenshot of checklist shown when exporting a conversation.}
    \label{fig:export-ui}
\end{figure}

\begin{figure*}[t]
    \centering
    \includegraphics[width=\textwidth]
    {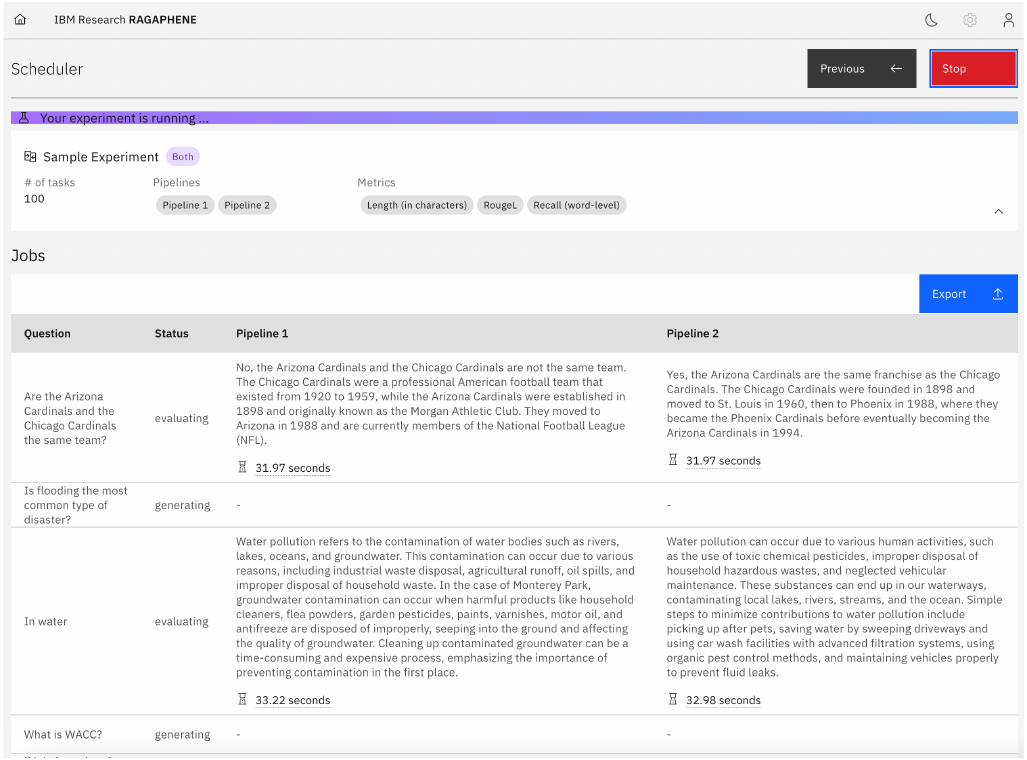}
    \caption{Screenshot of \workbench's experiment mode showing an experiment in progress.}
    \label{fig:experiment-ui}
\end{figure*}

We provide additional screenshots to showcase the platform features such as the comment functionality in Review mode (Figure~\ref{fig:comment-ui}), the checklist shown when exporting a conversation (Figure~\ref{fig:export-ui}), and running experiments using \workbench's Experiment mode (Figure~\ref{fig:experiment-ui}).

\end{document}